\documentclass[conference]{IEEEtran}
\IEEEoverridecommandlockouts
\usepackage{cite}
\usepackage{amsmath,amssymb,amsfonts}
\usepackage{color}
\usepackage{amsmath}
\usepackage{amssymb}
\usepackage{tabularx}

\usepackage{epsfig}
\usepackage{url}  

\usepackage{etoolbox}
\apptocmd{\sloppy}{\hbadness 10000\relax}{}{}

\usepackage{longtable}
\usepackage{graphicx}
\usepackage[T1]{fontenc}
\usepackage{paralist}
\usepackage{enumitem}
\usepackage{adjustbox}
\usepackage{array}
\usepackage{booktabs}
\newcolumntype{C}{>{\centering\arraybackslash}X} 
\usepackage{multirow}

\newcolumntype{R}[2]{%
    >{\adjustbox{angle=#1,lap=\width-(#2)}\bgroup}%
    l%
    <{\egroup}%
}
\newcommand*\rot{\multicolumn{1}{R{45}{1em}}}

\usepackage{epstopdf}

\def\BibTeX{{\rm B\kern-.05em{\sc i\kern-.025em b}\kern-.08em
    T\kern-.1667em\lower.7ex\hbox{E}\kern-.125emX}}

\usepackage{pdflscape}
\usepackage{tabularx}
\usepackage{xtab}

\usepackage{booktabs, threeparttable}
\usepackage{array}
\newcolumntype{L}[1]{>{\raggedright\arraybackslash}p{#1}}

\usepackage{algorithm}
\usepackage[noend]{algpseudocode}

\makeatletter
\def\BState{\State\hskip-\ALG@thistlm}
\makeatother

\algnewcommand\algorithmicswitch{\textbf{switch}}
\algnewcommand\algorithmiccase{\textbf{case}}
\algnewcommand\algorithmicassert{\texttt{assert}}
\algnewcommand\Assert[1]{\State \algorithmicassert(#1)}%
\algdef{SE}[SWITCH]{Switch}{EndSwitch}[1]{\algorithmicswitch\ #1\ \algorithmicdo}{\algorithmicend\ \algorithmicswitch}%
\algdef{SE}[CASE]{Case}{EndCase}[1]{\algorithmiccase\ #1}{\algorithmicend\ \algorithmiccase}%
\algtext*{EndSwitch}%
\algtext*{EndCase}%

\begin{document}


\title{A reinforcement learning approach to improve communication performance and energy utilization in fog-based IoT}

\author{\IEEEauthorblockN{Babatunji Omoniwa, Maxime Gu\'{e}riau and Ivana Dusparic}
\IEEEauthorblockA{\textit{School of Computer Science and Statistics} \\
\textit{Trinity College Dublin}\\
Dublin, Ireland\\
omoniwab@tcd.ie, \{maxime.gueriau, ivana.dusparic\}@scss.tcd.ie}
}

\maketitle

\begin{abstract}

Recent research has shown the potential of using available mobile fog devices (such as smartphones, drones, domestic and industrial robots) as relays to minimize communication outages between sensors and destination devices, where localized Internet-of-Things services (e.g., manufacturing process control, health and security monitoring) are delivered. However, these mobile relays deplete energy when they move and transmit to distant destinations. As such, power-control mechanisms and intelligent mobility of the relay devices are critical in improving communication performance and energy utilization. In this paper, we propose a Q-learning-based decentralized approach where each mobile fog relay agent (MFRA) is controlled by an autonomous agent which uses reinforcement learning to simultaneously improve communication performance and energy utilization. Each autonomous agent learns based on the feedback from the destination and its own energy levels whether to remain active and forward the message, or become passive for that transmission phase. We evaluate the approach by comparing with the centralized approach, and observe that with lesser number of MFRAs, our approach is able to ensure reliable delivery of data and reduce overall energy cost by 56.76\% -- 88.03\%.

\end{abstract}


\section{Introduction}
\begin{figure}[!t]
\centering
\includegraphics[width=2.6in]{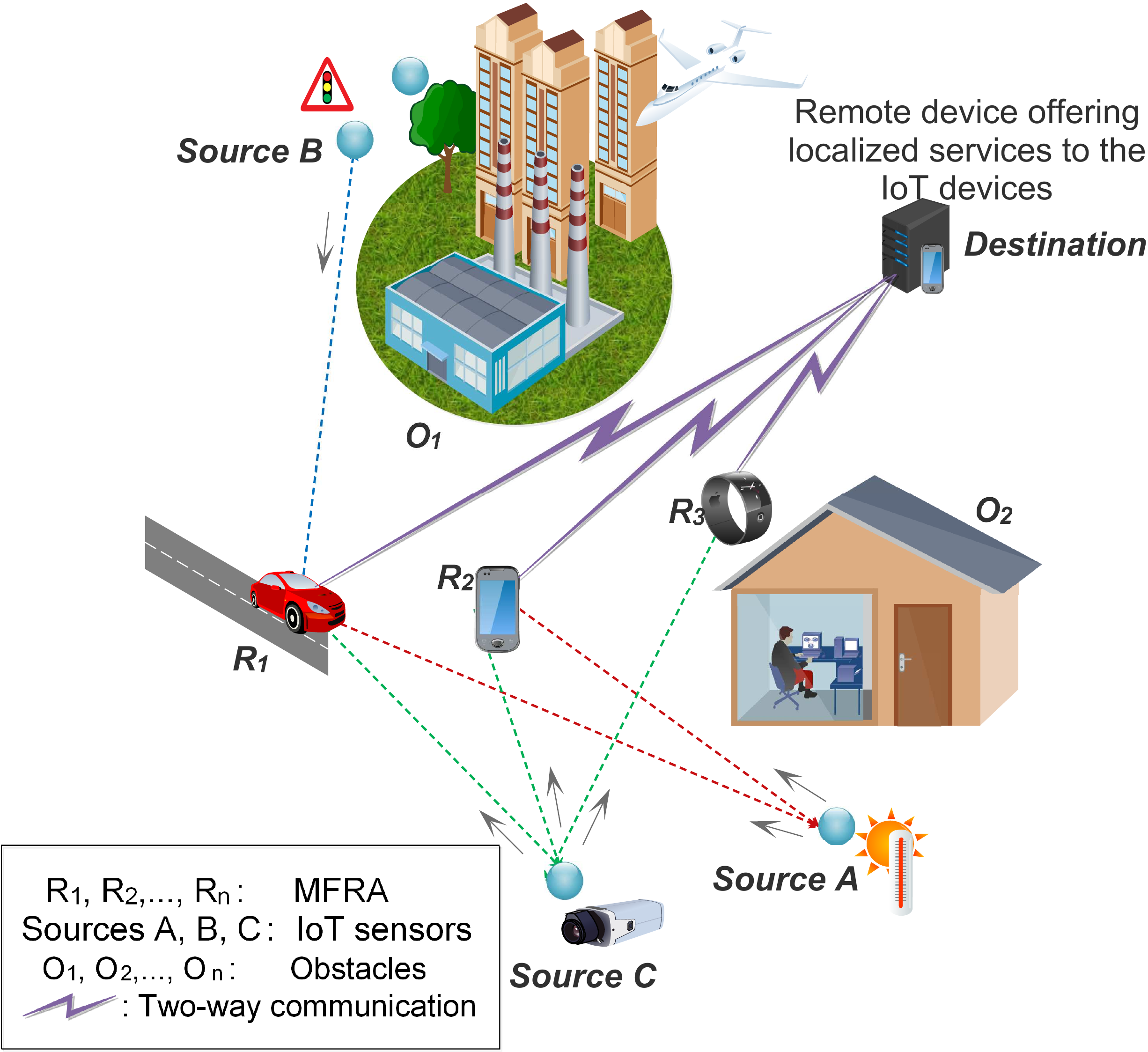}

\caption{The use of mobile relays in forwarding IoT traffic from source to destination.}
\label{ideafig1}
\end{figure}

\IEEEPARstart{R}{ecently}, the fog/edge computing-based Internet-of-Things paradigm was introduced to move computation, control, and decision-making from centralized data centers and IP backbone networks closer to IoT end-users at the network edge~\cite{Omoniwa2018}. Despite the proximity of ``things'' to local service providers at the edge of the network, communication outages may occur due to obstacles or long distances between a source (IoT sensor) and a remote destination node (where some IoT services are delivered) as illustrated in Fig. \ref{ideafig1}. The use of mobile fog devices as relays to forward IoT traffic between distant sources and destinations is shown to improve the overall network performance~\cite{Chiangh2016},~\cite{BenMimoune2017}. However, these relays are often power-constrained and deplete energy when they move and transmit to distant destinations. As such, power-control mechanisms and intelligent mobility of the relay devices are critical in improving communication performance and energy utilization.

\begin{table*}
 \footnotesize
\centering
\caption{Classification of Problems Addressed in Related Work}
\label{table:classification}
  \begin{tabular}{l|lllllllllll}
  \hline
 \textit{Reference} & \rot{\textit{Energy}} & \rot{\textit{Outage}} & \rot{\textit{Selection}} & \rot{\textit{Mobility}} & \rot{\textit{Latency}} & \rot{\textit{Traffic}} & \rot{\textit{Sources}} & \rot{\textit{Relays}} & \rot{\textit{Destination}} & \textit{Objective} & \textit{Approach} \\
  \hline \hline
   Omoniwa \emph{et al.}~\cite{OmoniwaRelay2018} & - & \checkmark & \checkmark & \checkmark & - & - & 1 & >1 & 1 & \multicolumn{1}{m{3cm}}{QoS} & \multicolumn{1}{m{3.2cm}}{Steepest Descent method} \\ \hline
   Simiscuka \emph{et al.}~\cite{Simiscuka2018} & - & \checkmark & \checkmark  & - & \checkmark & - & >1 & >1 & >1 & \multicolumn{1}{m{3cm}}{QoS} & \multicolumn{1}{m{3.2cm}}{Re-clustering Algorithm}\\ \hline
   Alsharoa \emph{et al.}~\cite{Alsharoa2018} & \checkmark & - & - & - & - & - & >1 & >1 & 1 & \multicolumn{1}{m{3cm}}{Energy, Planning} & \multicolumn{1}{m{3.2cm}}{Genetic Algorithm}\\ \hline
   Manzoor \emph{et al.}~\cite{Manzoor2018} & - & - & \checkmark & \checkmark & - & - & >1 & 1 & 1 & \multicolumn{1}{m{3cm}}{Relay selection} & Prototype design \\ \hline
    Lv \emph{et al.}~\cite{Lv2018} & \checkmark  & - & - & - & - & - & 1 & >1 & >1 & \multicolumn{1}{m{3cm}}{Energy} & Numerical \\ \hline
    Kawabata \emph{et al.}~\cite{Kawabata2017}  & - & \checkmark & \checkmark & - & - & - & >1 & >1 & >1 & QoS & \multicolumn{1}{m{3.2cm}}{Stochastic geometry}\\ \hline
    Behdad \emph{et al.}~\cite{Behdad2018} &  \checkmark & - & - & - & \checkmark & \checkmark & 1 & 1 & 1 & Energy, Congestion & Analytical \\ \hline 
     Hribar \emph{et al.}~\cite{Hribar2019} &  \checkmark & - & - & - & - & \checkmark & >1 & - & 1 & Energy & Deep Q-learning \\ \hline
      Wilhelmi \emph{et al.}~\cite{Wilhelmi2017} &  - & \checkmark & \checkmark & - & - & - & >1 & - & 1 & Aggregate throughput & Stateless Q-learning \\ \hline
       Azari \emph{et al.}~\cite{Azari2018} &  \checkmark &  \checkmark & - & - & - & - & >1 & - & 1 & QoS, Energy & Multi-armed bandit \\ \hline
     Our approach & \checkmark & \checkmark & \checkmark & \checkmark & - & - & >1 & >1 & >1 & QoS, Energy & \multicolumn{1}{m{3.2cm}}{Decentralized Q-learning} \\
      \hline \hline
 \end{tabular}
 \end{table*}

Several relay-based research have tried to address the issue of communication outages~\cite{OmoniwaRelay2018},~\cite{Simiscuka2018},~\cite{Kawabata2017}, energy usage~\cite{Alsharoa2018},~\cite{Lv2018},~\cite{Behdad2018}, relay selection~\cite{OmoniwaRelay2018},~\cite{Simiscuka2018},~\cite{Kawabata2017},~\cite{Manzoor2018}, mobility~\cite{OmoniwaRelay2018},~\cite{Manzoor2018}, latency~\cite{Simiscuka2018},~\cite{Behdad2018} and congestion~\cite{Behdad2018} within the IoT ecosystem. However, these research considered centralized approaches which are prone to several challenges. Such challenges include scalability, failure or downtime in the central entity. In addition, the overhead resulting from periodic updates and synchronization of nodes with the central controller often leads to inefficient energy utilization and decreased communication performance within the network. We further discuss these issues in Section II.

Several approaches have utilized learning to address communication outage and energy utilization, in order to allow devices to learn the optimal behaviours rather than being predefined. Reinforcement learning (RL) is an unsupervised learning technique in which an agent learns optimal behaviour through interaction with the environment~\cite{Sutton1998}. In \cite{Hribar2019}, deep Q-learning was used to prolong the lifetime of IoT sensors by improving energy utilization. Moreover, the RL-based approach does not require a centralized entity to control agents since each agent can be fully autonomous and take actions based on local information to improve overall system performance~\cite{Gueriau2018}~--~\cite{Azari2018}. In~\cite{Hribar2019} and~\cite{Azari2018}, a direct communication link between the IoT sensors and destination was considered which may not exist in reality due to obstacles and distance, as illustrated in Fig. \ref{ideafig1}.

In this paper, we take into account the distance and obstacles between source and destination, and leverage available mobile fog relays to simultaneously improve communication performance and energy utilization within the fog-based IoT network. The main contributions of this paper are as follows:
\begin{enumerate}
  \item A fully-decentralised reinforcement learning approach where each autonomous mobile fog relay agent (MFRA) learns to minimize communication outage within a fog-based IoT architecture.

  \item Taking into account mobility and death of MRFA (i.e. when the MFRA runs out of battery power) over time, we introduce a feedback mechanism where each destination device sends observations to neighbouring MFRAs. Based on local observation and feedback from the destination, each MFRA learns to either forward the message or become passive for a transmission phase, when it learns that there is a better performing (e.g., with higher probability of delivered packets) MFRA in its neighbourhood likely to forward the message instead. In this way, MFRAs effectively perform self-selection on which relay to forward messages at each transmission cycle, rather than the relay selection being done centrally. This strategy improves energy utilization within the network.
\end{enumerate}


\section{Related Work}
In this section, we present related work that address some challenges within the IoT domain. Table \ref{table:classification} highlights the most closely related work, addressing some of the key challenges within a relay-based IoT network. In~\cite{OmoniwaRelay2018}, the Steepest Descent method was used to minimize the communication outage in a fog-based IoT architecture. A mobile fog device is selected to serve as a relay from randomly distributed relaying candidates. However, the selection of an active relay is done centrally. In~\cite{Simiscuka2018}, a centralized cloud-deployed platform is responsible for selecting an optimal IoT smart gateway/relay node to improve network QoS. However, these research~\cite{OmoniwaRelay2018},~\cite{Simiscuka2018} did not consider energy efficiency of these power-constrained mobile relays.

An energy-efficient relaying scheme for IoT communications was presented in \cite{Alsharoa2018} to minimize energy consumption within an IoT network by solving the relay planning and QoS problems. A genetic algorithm-based approach was used to arrive at a near-optimal low-complexity solution. Using numerical methods, the authors in \cite{Lv2018} considered the energy-efficient design of a relay-based IoT network, in which multiple sources simultaneously transmit their information to the corresponding destinations via a relay equipped with a large array antenna. In~\cite{Behdad2018}, an analytical approach to maximize network effective lifetime by improving energy utilization was proposed. However, the relays considered in these approaches~\cite{Alsharoa2018},~\cite{Lv2018},~\cite{Behdad2018} are static and may not be suitable in a dynamic IoT environment.

The authors in~\cite{Manzoor2018} designed a prototype, a mobile relay architecture for low-power IoT devices, which exploits third-party unknown mobile relays for the forwarding of medical data generated by Bluetooth low energy sensors to some central server in the cloud. In~\cite{Kawabata2017}, a relay selection scheme for large-scale energy-harvesting IoT networks was proposed. The work applied a stochastic geometric approach and attempts to minimize the outage probability using a novel energy harvesting relay selection scheme. These approaches assume a centralized entity that has global knowledge of the network.

However, related work~\cite{OmoniwaRelay2018}~--~\cite{Behdad2018} mostly employ centralized approaches, which may not be suitable within the IoT domain due to several challenges, i.e. operational failure in central controller, management complexities, and increased signaling overhead.
Several approaches~\cite{Hribar2019},~\cite{Wilhelmi2017},~\cite{Azari2018} have utilized learning to address communication outage and energy utilization, in order to allow devices to learn the optimal behaviours rather than being predefined. In~\cite{Hribar2019}, deep Q-learning was used to prolong the lifetime of IoT sensors. A stateless variation of the Q-learning algorithm was presented in~\cite{Wilhelmi2017} to improve aggregate throughput of four coexisting network agents, however, this approach may not be adequate to solve more complex problems with larger state space. Another RL-based approach~\cite{Azari2018} proposed a distributed learning solution (a multi-armed bandit approach) against the centralized coordination. The work was able to achieve significant improvement in probability of success in data transmission and energy utilization of devices. However, a direct communication link between the IoT sensors and destination was considered in~\cite{Hribar2019} and~\cite{Azari2018}, which may not exist in reality due to obstacles and distance, as illustrated in Fig. \ref{ideafig1}.

With the possibility of leveraging available MFRs within the network, the aforementioned problems of relying on a central controller or assuming a direct line-of-sight communication link between source and destination can be well addressed. To this end, we present a decentralised reinforcement learning approach where each autonomous mobile fog relay agent (MFRA) uses local information to learn to simultaneously improve communication performance and energy utilization within a fog-based IoT architecture.

\section{Environment Model}
In this section, we describe the system model. We consider a network topology with randomly deployed low-cost IoT sensors having sensed data to transmit to remote destinations. However, due to long distances or obstacles between source and destination nodes, we consider a scenario where available mobile fog relay (MFR) nodes are used to forward these data/service request from IoT sensor to destination devices, thus, minimizing communication outage. However, MFRs energy-constrained and deplete energy when they move and transmit to distant destinations.

We assume the scenario presented in Fig. \ref{ideafig1} in Section I, with~$x$ number of sources~$(A, B, C, ...,~x)$, with~$x$ number of destinations, and the number of relays varying from 1~--~$x$. We observe that source~$A$ can reach the destination (local server) via two MFRs, source~$B$ via one MFR, and source~$C$ via three MFRs. For instance, when source~$A$ has data to send to the local server, it broadcasts a beacon via the wireless medium to all reachable neighbouring MFRs,~$N_{A} \in N$, where~$N_{A}$ denotes the set of all relays in the neighbourhood of source~$A$ i.e.~$N_{A} = \{R_1, R_2\}$   and~$N$ denotes all available MFRs in the network i.e.~$N= \{R_1, R_2 ..., R_n\}$. We assume that each MFR,~$R_{i} \in N$ is capable of forwarding a message from a single source during a transmission phase. When~$N_{A}$ receives the message from source~$A$, it forwards same message to the destination. In a centralized approach, the central controller is assumed to have global knowledge of all network devices and selects the best performing relay based on some pre-defined quality-of-service (QoS) metrics at the time of evaluation~\cite{Hribar2019}. However, we propose a decentralized approach where the local server broadcasts a feedback to all its neighbours,~$N^{D} \in N$, where~$N^{D}$ denotes the set of all MFRs in the neighbourhood of the destination~$D$ i.e.~$N^{D} = \{R_1, R_2, R_3\}$. Since~$R_3 \not\in N_{A}$, it discards the feedback message and remains passive for that transmission phase.

We model the outage behaviour from the environment using Equation (\ref{eqn1a}) as in~\cite{OmoniwaRelay2018}, which gives an estimate of the communication outage when the agent takes an action.
\begin{equation}\label{eqn1a}
  \mathcal{P}_{out} = 1 - (1 + 2\Psi^2 \ln \Psi) \exp\Big( -\frac{N_0 \tilde{\kappa}}{P_{I} (D_{I} + \delta)^{-\sigma}}\Big),
\end{equation}
where we define~$\Psi = \sqrt{(N_0 \tilde{\kappa})/(P_{R}(D_{S} + \delta)^{-\sigma})}$. The outage probability~$\mathcal{P}_{out}$ is defined as the probability that the signal-to-noise ratio falls below a predefined threshold~$\tilde{\kappa}$.~$P_{I}$ is transmit power of the IoT sensor,~$P_{R}$ is transmit power of the fog relay agent,~$D_{I}$ is the distance between IoT sensor and fog relay agent, and~$D_{S}$ is the distance between fog relay agent and destination node.~$\delta$ denotes the small positional change moved by the fog relay,~$N_0$ is the channel noise power, and~$\sigma$ is the path-loss exponent.

\section{Mobile Fog Relay Agent (MFRA)}
In this section, we present the reinforcement learning (RL) approach and briefly discuss the MFRA's states, actions and reward. In our scenario, the agents are the available mobile fog relays. The goal of the agent first, is to be alive during the transmission phase. Second, to relay message received from source to destination at a reasonably QoS by ensuring that the packets received in each transmission phase do not fall below some pre-defined threshold, which is set at 95\%. Finally, endeavour to be active when there are no potential relays to convey same message from IoT sensor to remote destination. An agent observes the environment locally (the energy consumed by itself) and from feedback updates from the destination device (the communication performance and availability of potential neighbours forwarding same traffic), which is explained in more details in this section.

\subsection{Reinforcement Learning Approach}
We apply the Q-Learning algorithm, an RL approach similar to~\cite{Gueriau2018} which requires no prior knowledge of the environment by the agent. In Q-learning, the agent interacts with the environment over periods of time according to a policy~$\omega$. At every time-step~$k \in K$, the environment produces an observation~$s_{k} \in \mathbb{R}^{D_s}$. By sampling, the agent then picks an action~$a_{k}$ over~$\omega(s_{k})$,~$a_{k} \in \mathbb{R}^{D_a}$, which is applied to the environment. The environment consequently produces a reward~$\mathcal{R}(s_{k}, a_{k})$ and may end the episode at state~$s_{K}$ or transits to a new state~$s_{k + 1}$. The agent's goal is to maximize the expected cumulative reward,~$\max_{\omega} \mathbb{E}_{s_0, a_0, s_1, a_1, ..., s_K} \Big[ \sum_{i=0}^{K} \gamma^i \mathcal{R}_{k + i + 1} \Big]$, where~$0 \leq \gamma \leq 1$ is the discount factor, and~$\mathcal{R}_{k + i + 1}$ is the reward received at each state.

First, the agent takes an initial random action~$a_{k}$ and gets observations from the environment which corresponds to that action, as well as a reward. The agent then updates it's Q-values at each time-step~$k$ following Equation (\ref{eqn1b}).

\begin{equation}\label{eqn1b}
\begin{split}
Q(s_k, a_k) &:= Q(s_k, a_k)\\
& + \alpha \Big[ \mathcal{R}_{k + 1} + \gamma \max_{a}  Q(s_{k + 1}, a) -  Q(s_k, a_k) \Big],
   \end{split}
\end{equation}
where~$\alpha$ is the learning rate, which determines the impact of new experience on the Q-value,~$\mathcal{R}_{k + 1}$ is the reward the agent receives by being in~$s_{k + 1}$ from~$s_{k}$. Based on the policy followed by the agent, it gets observations and rewards from the environment.

\subsection{Agent Design}
The agent's states, actions and reward are explained below.

\emph{States}: The states are defined as a tuple, $\langle$Outage communication cost ($\mathcal{P}_{out}$) /Energy status of the fog relay /Neighbour potential to relay message (Availability of redundant nodes)$\rangle$. The agent then discretize the continuous observations emanating from the environment into a~$3 \times 3 \times 3$ state space.

\begin{itemize}
  \item \emph{Outage communication cost}: Outage observations from the environment is estimated using (\ref{eqn1a}) from \cite{OmoniwaRelay2018}, which gives an estimate of the communication outage when the agent takes an action, such as move and transmit. The agent discretizes this observation into 3 states: 0\% -- 33\%, 34\% -- 66\%, and 67\% -- 100\%.
  \item \emph{Energy status of the fog relay}: This observation gives the agent insight on how much energy is consumed by the fog agent when following policy~$\omega_i \in \omega_{fog}$. If the fog agent continues to take sub-optimal actions, it depletes its energy and dies out. The agent discretizes this observation into 3 states: 0\% -- 33\%, 34\% -- 66\%, and 67\% -- 100\%.
  \item \emph{Neighbour potential to relay message (Availability of redundant nodes)}: This observation gives the agent insight on the availability of redundant nodes that can help in relaying same type of message emanating from a particular IoT sensor. If there are no potential relay agent to convey message from an IoT sensor to a remote destination, then the agent should learn to remain active for that transmission phase. However, if there are one or more potential relays agents, the agent should learn to take no action to help conserve energy and improve the longevity of the network. The agent discretizes this observation into 3 states: no available redundant relay, 1 available redundant relay, and more than one available redundant relay.
\end{itemize}

\emph{Action space}: The actions for the fog relay agent are move closer ($-\delta$) and transmit, move farther ($+\delta$) and transmit, and do nothing (become passive). An action is executed in each step until the end of an episode.

\emph{Goal}: The agent's goal is to be alive during the transmission phase and ensure that packets received in each transmission phase do not fall below some pre-defined threshold, which is set at 95\%.

\emph{Reward $\mathcal{R}$}: The reward function used is given in Equation~(\ref{eqn1c}) as

 \begin{equation}\label{eqn1c}
    \mathcal{R} =
    \begin{cases}
      100, & \text{if}\ \emph{goal}~is~Reached \\
      0, & \text{otherwise.}
    \end{cases}
  \end{equation}

The MFRA's learning process is summarized in Algorithm \ref{fiotRL}. A learning episode is terminated when the agent attains the pre-defined goal or when the MFRA dies, possibly due to taking sub-optimal actions without reaching the goal, or when the maximum step for an episode is reached. Intuitively, an MFRA should take an action of doing nothing if there are other available agents more capable of transmitting during that transmission phase, so as to conserve energy. On the contrary, if the fog relay agent continues to move and transmit even when there is sufficient redundancy within the network to relay same message, it may die out sooner, therefore causing a point-of-failure in the network.

\begin{algorithm}
\footnotesize
\caption{MFRA Learning Process}\label{fiotRL}
\begin{algorithmic}[1]
\State \textbf{Initialize:}~maxStep = 100000
\For{Episodes = 1, 2, 3, ...}
\State \texttt{\%\% An episode ends when \emph{goal} is Reached or Agent \emph{dies} or maxStep is reached}
\State ResetEnvironment()
\While{\emph{goal} not Reached and Agent \emph{alive} and maxStep not reached}
\State \emph{state} $\leftarrow$ MapLocalObservationToState(\emph{Env})
\State \emph{action} $\leftarrow$ QLearning.SelectAction(\emph{state})
\State \texttt{\%\% AgentExecutesActionInState}
\State  \emph{action}.execute(\emph{Env})
\If { \emph{action}  == ``move close and Tx'' or ``move away and Tx''} 
\State \texttt{\%\% MapToNewState}
\State Env.EstimateOutage using (\ref{eqn1a})
\State Env.EstimateFogEnergyUsed
\State Env.CheckAvailableActiveFogNeighbour
\State \emph{state} $\leftarrow$ MapObservationToState(\emph{Env})
\ElsIf {\emph{action}  == ``do nothing''} 
\State \texttt{\%\% MapToNewState}
\State Env.Outage $\leftarrow$ 100\%
\State Env.FogEnergyUsed $\leftarrow$ 0\%
\State Env.CheckAvailableActiveFogNeighbour
\State \emph{state} $\leftarrow$ MapObservationToState(\emph{Env})
\EndIf
\State \textbf{endif}
\State UpdateQLearningProcedure() (\ref{eqn1b})
\State EvaluateReward() (\ref{eqn1c})
\EndWhile
\State \textbf{endwhile}
\EndFor
\State \textbf{endfor}
\end{algorithmic}
\end{algorithm}

\section{Simulation Setup}

\begin{table}
\footnotesize
\centering
\caption{Simulation Parameters}
\label{table:simparameters}
\begin{tabular}{lll}
  \hline
 &\textit{Parameter} & \textit{Values} \\
  \hline \hline
   \parbox[t]{2mm}{\multirow{7}{*}{\rotatebox[origin=c]{90}{Environment}}} &Simulation space & $80 \times 80$ metres\\
      &$P_{I}$ & [0.001, 0.3] Watts \\
   &$P_{R}$ & 0.3 Watts\\
  & $\delta$ & $\pm0.25$ metres\\
   &MFRA mobility bound & [-30,~30] metres\\
   &Noise power $N_0$ & $2 \times 10^{-7}$ Watts\\
  & Path-loss exponent $\sigma$ & 3\\
  & Predefined threshold $\tilde{\kappa}$ & 1\\ \hline
  \parbox[t]{2mm}{\multirow{5}{*}{\rotatebox[origin=c]{90}{Agent}}} & Discount factor $\gamma$ & 0.9\\
  & Learning rate $\alpha$ & 0.1\\
  & Episodes $N$ & 100\\
   &Maximum iteration runs & 100000\\
  & Policy $\epsilon$ & $e^{-0.0015N}$\\

   \hline \hline
 \end{tabular}
 \end{table}

In this section, we present the simulation setup. We carry out experiments using Python to evaluate the performance of the proposed Q-learning approach. The simulation parameters are summarized in Table~\ref{table:simparameters}. In our scenario, we set each mobile fog relay (MFR) with a degree of 6, this implies that each MFR can have a maximum of 6 neighbouring sources (IoT sensors). In each transmission phase, any of the 6 IoT sensors can transmit sensed data from its environment to neighbouring MFRs~$N_{A} \in N$ within its communication range, and each MFR is capable of receiving data from a single source~$A$ and forward the data to a destination device. It is possible for two or more MFRs, which are neighbours~$N_{A}$, to receive same message from a single source~$A$ as depicted in Fig. \ref{ideafig1}.

We compare the performance of the proposed RL-based approach with a baseline centralised approach based on~\cite{OmoniwaRelay2018},~\cite{Simiscuka2018}. In a centralized approach, the central controller is assumed to have global knowledge of the entire network and selects a potential MFR that satisfies the highest QoS requirement (the communication link via a MFR with minimum outage) to become an active relay for that transmission phase. For instance, when the destination device receives the message from~$N_{A}$, the network controller examines the number of successfully delivered packets and selects the best MFR based on the successful packets delivery. Rather than have selection done by the central entity, we introduce a feedback mechanism where each destination device sends observations (the percentage of successfully delivered packets and the availability of potential redundant MFR) to its neighbours~$N^{D}$. Based on this feedback and local observation (which gives the energy status of the MFRA), each MFRA performs self-selection and learns to become passive for a particular transmission phase when there is an active MFRA within its neighbourhood.

We apply the Q-learning algorithm using 100 learning episodes. An episode ends either when the MFRA reaches its goal, dies-out due to energy outage, or when the maximum step for an episode is reached. We evaluate the performance of the proposed approach using two metrics, namely:
\begin{itemize}
  \item Successful packets delivery: We evaluate this metric as the percentage of received packets at the destination node with respect to that transmitted by the IoT sensor via the active MFRA.
  \item Total energy consumption: We evaluate this metric as the percentage of total energy consumed with respect to the initial capacity of an active MFRA.
\end{itemize}

\section{Results}
\begin{figure}[!t]
\centering
\includegraphics[width=2.8in]{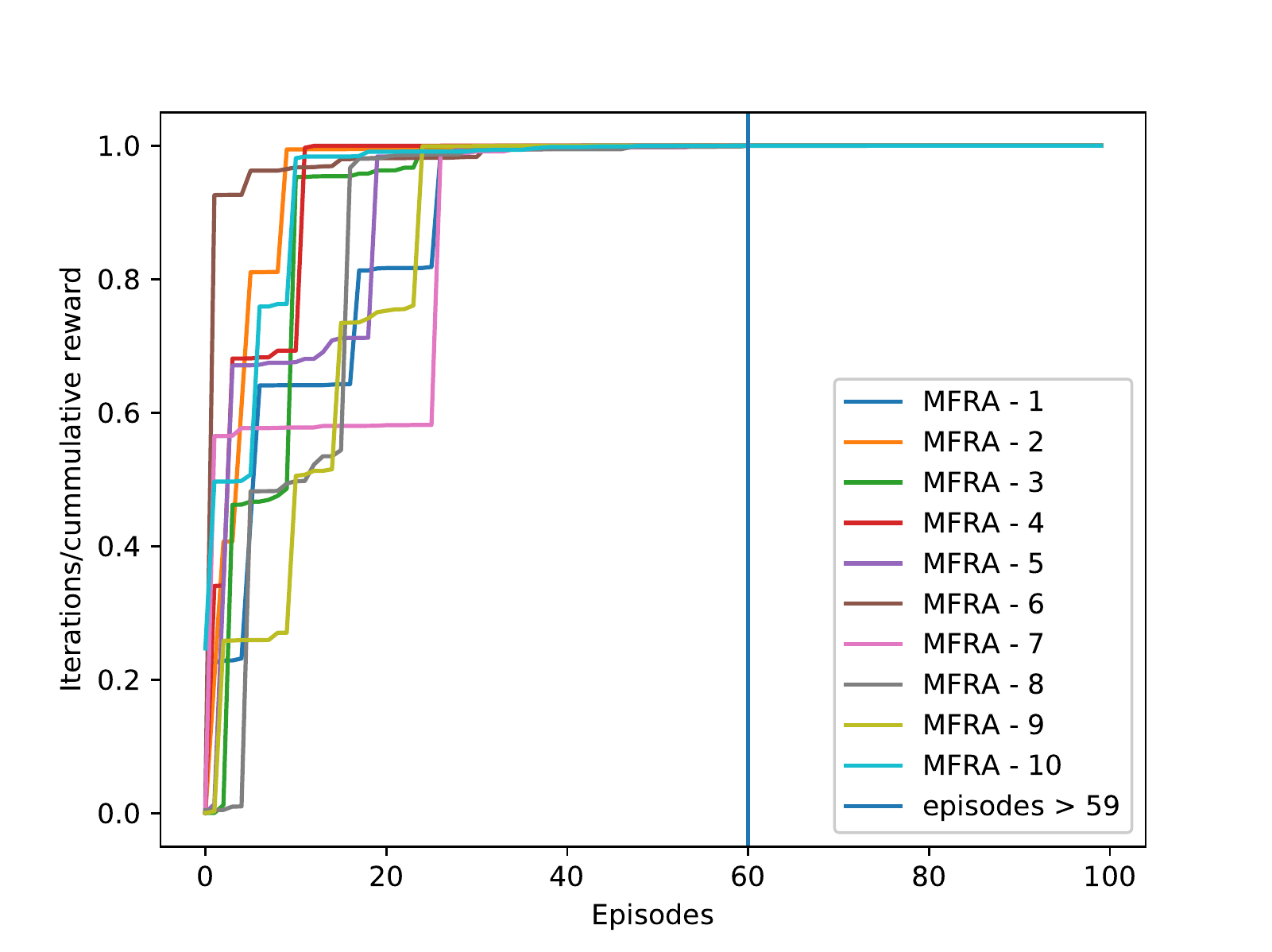}
\caption{Number of iterations per cumulative reward of each agent vs. the number of episodes.}
\label{cum_reward}
\end{figure}

In this section, we present experimental results and discussions. Fig. \ref{cum_reward} show the number of iterations per cumulative reward of each agent over 100 learning episodes. From the figure, convergence for each MFRA is achieved in about 50 episodes. Without loss of generality, we consider results from the last 40 episodes of the simulation since each agent is expected to have learnt to behave. Furthermore, we carry out 50 runs of experiments to validate our findings.

\begin{figure}[!t]
\centering
\includegraphics[width=2.6in]{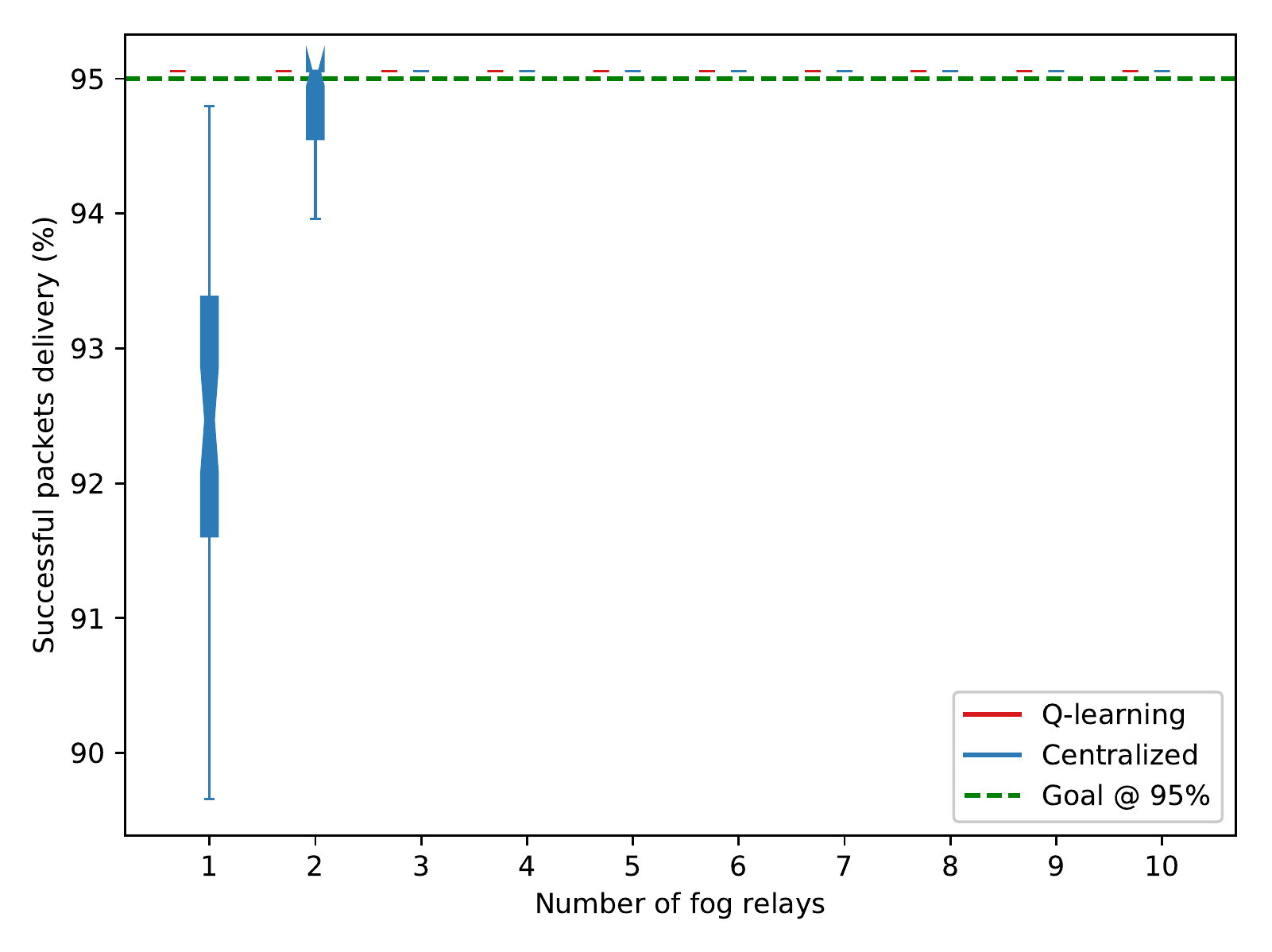}
\caption{Successful packets delivery vs. the number of fog relays.}
\label{outage}
\end{figure}

\begin{figure}[!t]
\centering
\includegraphics[width=2.6in]{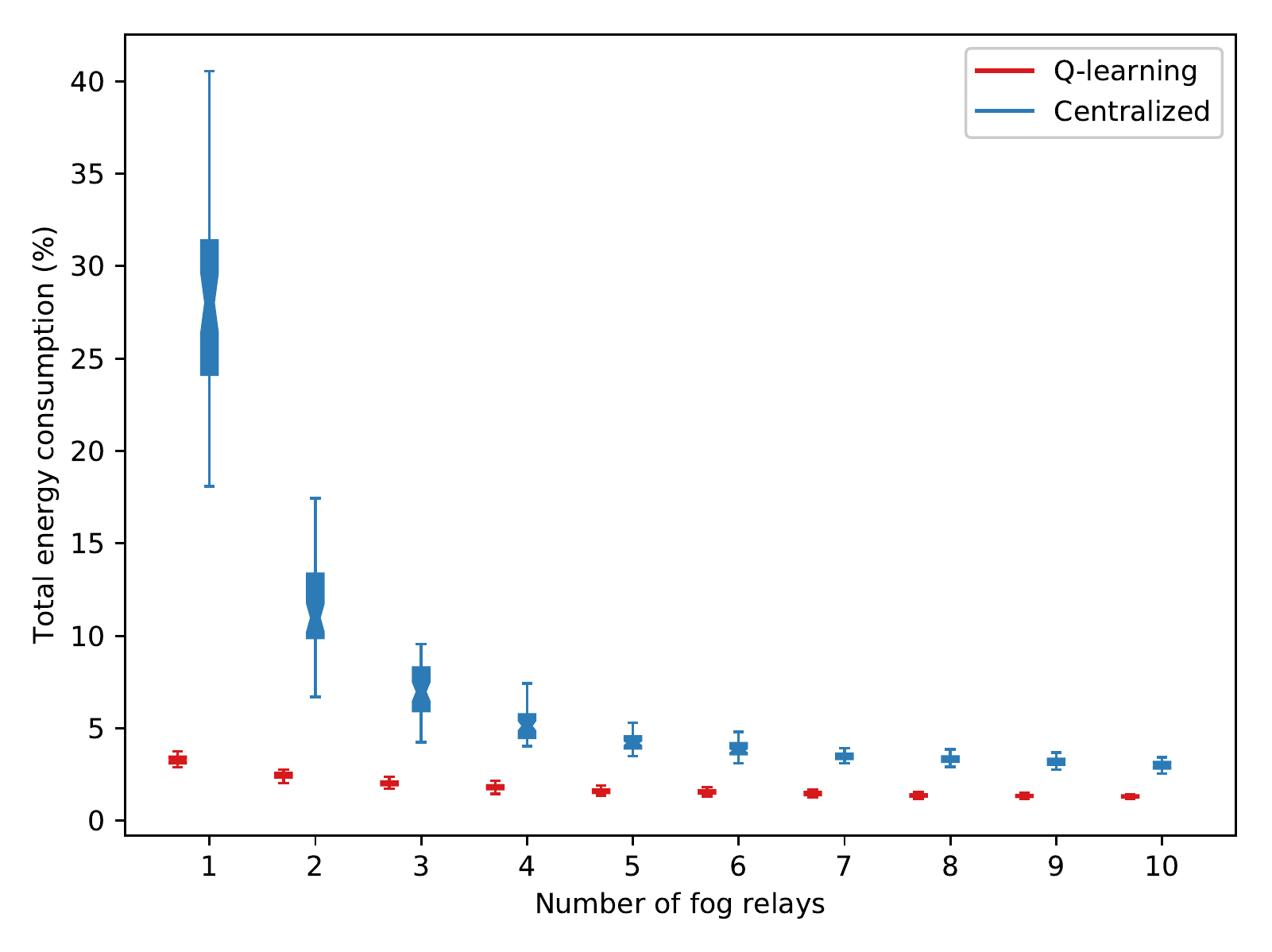}
\caption{Total energy consumption vs. the number of fog relays.}
\label{energy}
\end{figure}

Fig. \ref{outage} shows percentage of successfully delivered packets when different number of fog relays are used in the network. From the figure, the performance of both approaches was same when three or more mobile fog relays were used, however, the centralized approach performed poorly when compared with the proposed approach with two or less number of relays. Intuitively, when more redundant relays are deployed in the network, we may observe better performance, but at higher hardware cost. However, the poor performance in the centralized approach shows the limitation of the central controller in selection. Fig. \ref{energy} illustrates the energy usage of active mobile fog relays when the number of fog relays in the network are varied. Interestingly, we observe an exponential decay in the consumed energy for both approaches, however, the energy utilization of the proposed approach outperformed the centralized approach with an improvement of about 56.76\% -- 88.03\%. When more MFRs are deployed to the network, we observe lesser energy consumption. Intuitively, when there are available redundant relays within the network, lesser amount of energy is required by individual MFRs.

\begin{figure}[!t]
\centering
\includegraphics[width=2.6in]{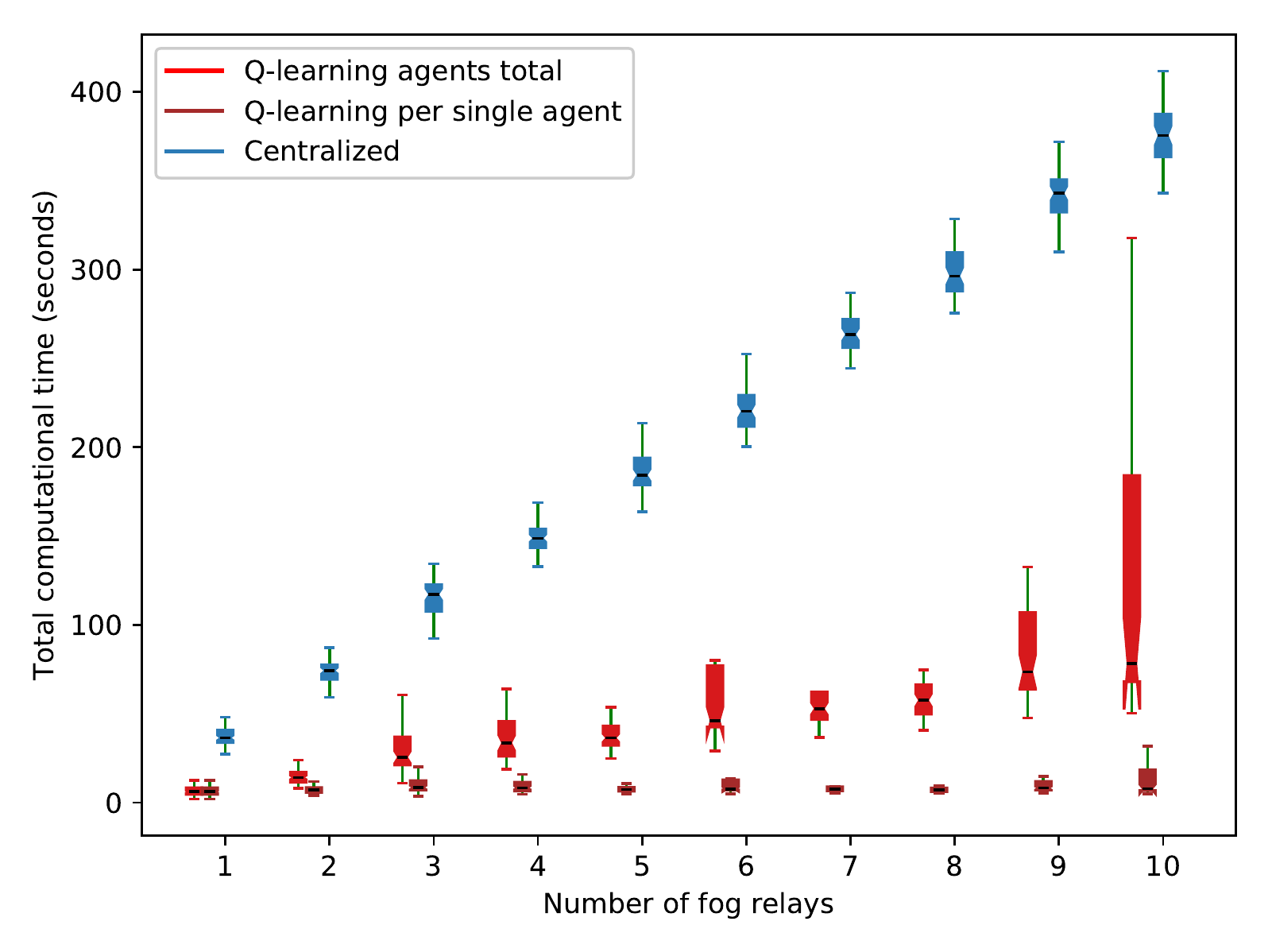}
\caption{Total computational time of different approaches vs. the number of fog relays.}
\label{time}
\end{figure}

In Fig. \ref{time}, we show the computational time it takes the mobile fog relay to complete all episodes. From the figure, we observe that both approaches exhibit linear growth in total computational time as the number of fog relays increased.
However, while the overall computation time increases, the decentralized approach effectively distributes computation across the MFRAs as seen in the Q-learning per single agent unlike the centralized approach. So in time-constrained scenarios, it is much quicker to achieve efficient energy utilization and successful packets delivery in the decentralized RL-based approach than the centralized approach.

\section{Conclusion and Future Work}
Communication outage within the IoT network can be minimized by leveraging available mobile fog relays (MFRs). However, these MFRs are power-constrained and deplete energy when they move and transmit to distant destinations. As such, power-control mechanisms and intelligent mobility of the relay devices are critical in improving communication performance and energy utilization. In this paper, we investigated a Q-learning-based decentralized approach where each autonomous mobile fog relay agent (MFRA) performs self-selection based on a feedback mechanism from the destination and use local information to simultaneously learn to improve communication performance and energy utilization. Comparative analysis was carried out with a baseline centralized approach. Results from simulations reveal that our proposed reinforcement learning (RL)-based approach achieved about 56.76\% -- 88.03\% improvement in energy utilization as compared to the baseline. We observe that energy improvement may be achieved in the centralized approach only when more redundant relays are utilized, which is not cost effective. Furthermore, with lesser number of MFRs, the centralized approach failed to ensure reliable data delivery, unlike the proposed RL-based approach. However, increasing the traffic emanating from the heterogenous IoT sensors may have some impact on the performance of the network.

One of the most interesting directions for future work is to evaluate the impact of learning on increased network traffic from heterogenous sources and take into consideration mission-critical IoT applications in order to meet stringent QoS requirements.

%

\section*{Acknowledgement}
This publication has emanated from research supported in part by a research grant from Science Foundation Ireland (SFI) under Grant Number 13/RC/2077 and 16/SP/3804 and by Irish Research Council through Surpass New Horizons award.

\end{document}